\pdfoutput=1
\documentclass[11pt,a4paper]{article}
\usepackage[hyphens]{url}
\usepackage[hyperref]{acl2021}
\usepackage{times}
\usepackage{latexsym}

\usepackage{multirow}
\usepackage{booktabs}
\usepackage{graphicx}
\usepackage{stfloats}
\usepackage{placeins}
\usepackage{amsmath}
\usepackage{amssymb}
\usepackage{bm}
\usepackage{paralist}
\usepackage{makecell}

\usepackage{xcolor}
\usepackage{soul}

\usepackage{microtype}

\aclfinalcopy 
\def\aclpaperid{1878} 

\definecolor{emotion}{RGB}{255, 234, 167}
\definecolor{event}{RGB}{199, 224, 235}

\newcommand{\emotion}[1]{
  \begingroup
  \sethlcolor{emotion}
  \emph{\hl{#1}}
  \endgroup
}

\newcommand{\event}[1]{
  \begingroup
  \sethlcolor{event}
  \textbf{\hl{#1}}
  \endgroup
}

\title{Stylized Story Generation with Style-Guided Planning}

\author{Xiangzhe Kong\Thanks{~Equal contribution.}, Jialiang Huang\footnotemark[1], Ziquan Tung, Jian Guan and Minlie Huang%
\Thanks{~Corresponding author.}\\
The CoAI group, DCST, Institute for Artificial Intelligence,\\
State Key Lab of Intelligent Technology and Systems,\\
Beijing National Research Center for Information Science and Technology,\\
Tsinghua University, Beijing 100084, China\\
\texttt{\{kxz18,huang-jl17,tongzq18,j-guan19\}@mails.tsinghua.edu.cn},\\ \texttt{aihuang@tsinghua.edu.cn} \\
}

\date{}

\begin{document}
\maketitle
\begin{abstract}
Current storytelling systems focus more on generating stories with coherent plots regardless of the narration style, which is important for controllable text generation.
Therefore, we propose a new task, stylized story generation, namely generating stories with specified style given a leading context. 
To tackle the problem, we propose a novel generation model that first plans the stylized keywords and then generates the whole story with the guidance of the keywords. Besides, we propose two automatic metrics to evaluate the consistency between the generated story and the specified style. Experiments demonstrates that our model can controllably generate \textit{emotion}-driven or \textit{event}-driven stories based on the ROCStories dataset \citep{mostafazadeh2016corpus}. Our study presents insights for stylized story generation in further research.
\end{abstract}

\section{Introduction}
\begin{figure}[htp]
    \centering
    \includegraphics[width=.5\textwidth]{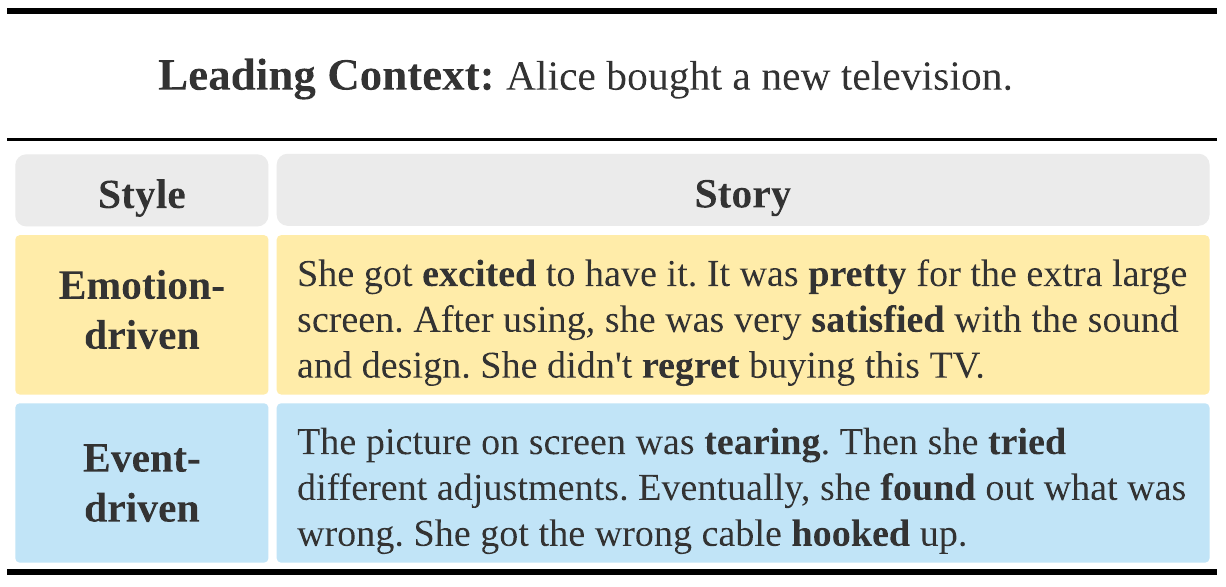}
    \caption{Example of stylized story generation given the same leading context. The stylized keywords are in bold.}
    \label{fig:stylized_text}
\end{figure}

Story generation is a challenging task in natural language generation (NLG), namely generating a reasonable story given a leading context.
Recent work focuses on 
enhancing the coherence of generated stories \citep{fan2018hierarchical, yao2019plan} or introducing commonsense knowledge~\cite{DBLP:journals/tacl/GuanHHZZ20,DBLP:conf/emnlp/XuPSPFAC20}.
However, 
it has not yet been investigated to generate stories with controllable styles, which is important since different styles serve different writing purposes. 
As exemplified in ~\autoref{fig:stylized_text}, \textit{emotion-driven} stories use emotional words~(e.g., \textit{``excited''}, \textit{``enjoyed''}) to reveal the inner states of the characters and bring the readers closer to the characters.
In comparison, \textit{event-driven} stories usually contain a sequence of events with a clear temporal order 
(e.g., \textit{``tearing''}$\rightarrow$\textit{``tried''}$\rightarrow$\textit{``found''}$\rightarrow$\textit{``hooked''}), which aims to narrate the story objectively.

In this paper, we formalize the task of stylized story generation, which requires generating a coherent story with a specified style given the first sentence as the leading context.
\textit{Style} has multiple interpretations, which can be seen as a unique voice of the author expressed through the use of certain stylistic devices (e.g. choices of words)\cite{mou-vechtomova-2020-stylized}.
In this work 
we focus on the choices of words and define the story styles based on the pattern of wording. Specifically, we focus on two story styles, including \textit{emotion-driven} and \textit{event-driven} stories.
\textit{Emotion-driven} stories contain abundant words with emotional inclination. We identify the
emotional words using the off-the-shelf toolkit NRCLex~\citep{DBLP:journals/corr/abs-2011-03492}, which supports retrieving the emotional effects of a word from a predefined lexicon. 
And \textit{event-driven} stories tend to use serial actions as an event sequence.
We use NLTK~\citep{bird2009natural} to extract verbs in a story as the actions.
Since no public datasets are available for learning to generate stylized stories, 
we regard the extracted words as stylistic keywords and then annotate the story styles for existing story datasets automatically based on the keyword distribution. 
Note that the story styles can be extended easily by defining new stylistic keywords. 

In this work, we propose a generation model 
for stylized story generation. 
Our model first predicts the distribution of stylistic keywords and then generates a story with the guidance of the distribution.
Furthermore, we propose two new automatic metrics to evaluate the consistency between the generated stories and the specified styles: lexical style consistency (LSC) and semantic style consistency (SSC), which focus on the number of stylistic keywords and the overall semantics, respectively. 
Extensive experiments demonstrate that  
the stories generated by our model not only achieve better fluency and coherence 
than strong baselines 
but also have better consistency with the specified styles.\footnote{Link to the code: \url{https://github.com/thu-coai/Stylized-Story-Generation-with-Style-Guided-Planning.git}}

\section{Related Work}
\paragraph{Story Generation}
Recently there have been significant advances for story generation with the encoder-decoder paradigm~\cite{sutskever2014sequence}, the transformer-based architecture~\cite{vaswani2017attention} and the large-scale pre-trained models~\cite{radford2019language,lewis2020bart}.
Prior studies usually decomposed the generation into separate steps by first planning a sketch and then generating the whole story from the sketch. The sketch is usually a series of keywords~\cite{yao2019plan}, a learnable skeleton~\cite{DBLP:conf/emnlp/XuRZZC018} or an action sequence~\cite{DBLP:conf/acl/FanLD19,DBLP:conf/emnlp/Goldfarb-Tarrant20}. Another line is to incorporate external knowledge into story generation~\cite{DBLP:journals/tacl/GuanHHZZ20, DBLP:conf/emnlp/XuPSPFAC20}. 
However, generating stories with controllable styles has hardly been investigated. 
\paragraph{Stylized Generation} 
Stylized generation aims to generate texts with controllable attributes.
For example, recent studies in dialogue systems focused on controlling persona~\cite{zhang2018personalizing,boyd2020large}, sentence functions~\cite{ke-etal-2018-generating}, politeness~\cite{TACL1424}, and topics~\cite{tang2019target}. In story generation, \citet{huang2019hierarchically} and \citet{DBLP:conf/emnlp/XuPSPFAC20} controlled the story topics and planned keywords, respectively. 
Besides, for general text generation, the authorship~\cite{tikhonov2018guess}, sentiment~\cite{hu2017toward}, and topics~\cite{ li2020complementary} can also be controlled for different purposes.
We introduce a new controllable attribute in story generation, i.e., the story style, which has been paid little attention to in prior studies.

\section{Proposed Method}
In this section, we first show the task formulation for stylized story generation~(\S{\ref{subsec:problem-formulation}}). Then we present the details of our two-step model: \textbf{style-guided keywords planning}~(\S{\ref{subsec:keywords-distribution}}) and \textbf{generation with planned keywords}~(\S{\ref{subsec:generation}}).

\subsection{Task Formulation} \label{subsec:problem-formulation}

\textbf{Input:} The first sentence $\bm{x} = (x_1, x_2, \ldots, x_n)$ of a story with length $n$, where $x_i$ is the $i$-th word. A special token $l$ to indicate the expected style of the generated story. $l \in \{\langle \texttt{emo}\rangle, \langle \texttt{eve}\rangle\}$, which refers to the \textit{emotion-driven} and \textit{event-driven} styles, respectively. Besides, in the training phase, we set $l=\langle \texttt{other}\rangle$ if the training example is neither {emotion-driven} nor {event-driven} to improve the data efficiency. 

\noindent\textbf{Output:} 
A story $\bm{y}=(y_1, y_2, \ldots, y_m)$ of length $m$ with the style $l$, where $y_i$ is $i$-th word. 

\subsection{Planning}
\label{subsec:keywords-distribution}
We insert $l$ at the beginning of $\bm{x}$ and encode them as follows:
\begin{align}
    [\bm{h}_0, \bm{h}_1, \ldots, \bm{h}_n]= \texttt{Enc}(l,x_1, x_2,\ldots,x_n),
\end{align}
where $\bm{h}_i~(1\leqslant i\leqslant n)$ is the hidden state corresponding to $x_i$, $\bm{h}_0$ is the hidden state at the position of $l$, and \texttt{Enc} is a bidirectional or unidirectional encoder.
Then, we regard the stylistic keywords as bag-of-words \cite{kang2020self} and predict the keyword distribution $P_{k}(w|\bm{x},l)$ over the whole vocabulary $\mathbb{V}$ as follows:
\begin{equation}
    P_{k}(w|\bm{x},l) =\text{softmax}(\textbf{W}_k\bm{h}_c + \textbf{b}_k),
\end{equation}
where $\textbf{W}_k$ and $\textbf{b}_k$ are trainable parameters, and $\bm{h}_c$ is the context embedding to summarize the input information. We directly set $\bm{h}_c=\bm{h}_0$. 
The training objective in this stage is to minimize the cross-entropy loss $\mathcal{L}_{k}$ between the predicted keyword distribution $P_{k}(w|l,\bm{x})$ and the ground truth $\hat{P}_{k}(w|l,\bm{x})$ as follows:
\begin{equation}\label{eq:keywords_loss}
    \mathcal{L}_{k} = -\sum_{i=1}^{|\mathbb{V}|} \hat{P}_{k}(w_i|l, \bm{x}) \log {P}_{k}(w_i|l, \bm{x}),
\end{equation}
where $w_i$ denotes the  $i$-th word in $\mathbb{V}$ and $\hat{P}_{k}(w|l,\bm{x})$ is an one-hot vector over $\mathbb{V}$.
We do not decode a keyword sequence explicitly~\cite{yao2019plan}
but generate stories directly based on the keyword distribution $P_k(w|l,\bm{x})$ to avoid introducing extra exposure bias~\cite{DBLP:journals/corr/abs-1905-10617}.


\subsection{Generation}
\label{subsec:generation}
We employ a left-to-right decoder to generate a story conditioned upon the input and the predicted keyword distribution. The training objective in this stage is to minimize the negative log-likelihood $\mathcal{L}_{st}$ of the ground truth stories:
\begin{align}
    \mathcal{L}_{st} &= -\sum_{t=1}^m\log {P}(y_t|l,\bm{x},y_{<t}).
\end{align}
We derive ${P}(y_t|l,\bm{x},y_{<t})$ by explicitly combining the stylistic keyword distribution into the decoding process 
as follows:
\begin{align}
{P}(y_t|l,\bm{x},y_{<t}&)= P_l(y_t|l,\bm{x},y_{<t})\cdot (1-\bm{g}_t)\notag\\&~~~~~~~~~~+P_k(y_t|l,\bm{x})\cdot \bm{g}_t,\\
    P_l(y_t|l,\bm{x},y_{<t}) &= \text{softmax}(\textbf{W}_s\bm{s}_t+\textbf{b}_s),\\
    \bm{s}_t &= \texttt{Dec}({y}_{<t}, \{\bm{h}_i\}_{i=0}^n),
\end{align}
where $\textbf{W}_s$ and $\textbf{b}_s$ are trainable parameters, $P_l$ is a 
distribution over $\mathbb{V}$ without conditioning on the predicted keywords, and $\bm{g}_t\in\mathbb{R}^{|\mathbb{V}|}$ is a gate vector indicating the weight of the keyword distribution $P_k$. We compute $\bm{g}_t$ as follows:
\begin{align}
    \bm{g}_t &= \text{sigmoid}(\textbf{W}_g [\bm{r}_t;\bm{s}_t] + \bm{b}_g),\\
    \bm{r}_t &= \bm{W}_r P_k(y_t | l, \bm{x}) + \bm{b}_r, 
\end{align}
where $\bm{W}_g$, $\bm{b}_g$, $\bm{W}_r$ and $\bm{b}_r$ are trainable parameters.
In summary, the final training objective $\mathcal{L}$ of our model is derived as follows:
\begin{equation}
    \label{eq:loss}
    \mathcal{L} = \mathcal{L}_{st} + \alpha \cdot \mathcal{L}_{k}.
\end{equation}
where $\alpha$ is an adjustable scale factor. 

\section{Experimental Setup}
  \subsection{Dataset} \label{subsec:dataset}
 We conduct the experiments on the ROCStories corpus \cite{mostafazadeh2016corpus}, which contains 98,159 five-sentence stories. We randomly split ROCStories by 8:1:1 for training/validation/test, respectively. The average number of words in the input~(the first sentence) and the output~(the last four sentences) are 9.1 and 40.8, respectively. Besides, we follow \citet{DBLP:journals/tacl/GuanHHZZ20} to delexicalize stories in the dataset by masking all the male /female/neutral names with $\langle \texttt{MALE}\rangle$/ $\langle \texttt{FEMALE}\rangle$/ $\langle \texttt{NEUTRAL}\rangle$ to achieve better generalization.
 
  \subsection{Style Annotation}
  We extract stylistic keywords from stories in the dataset and assign a style label for each story according to the distribution of stylistic keywords.\label{sec:anno}
    \paragraph{Stylistic Keywords} We use NRCLex and NLTK to extract stylistic keywords. 
    NRCLex maps each word in a story to its underlying emotion labels according to a 
    word-emotion lexicon~(e.g., \textit{``favorite"} $\rightarrow$ 
    \textit{``joy''}). We select the words with following emotion labels: \textit{``fear'', ``anger'', ``surprise'', ``sadness'', ``disgust''} and \textit{``joy''}, as the keywords for the {emotion-driven} style. Besides, we use NLTK to extract verbs as keywords for the {event-driven} style. We filter out the stop words and common verbs with bottom ten IDF\footnote{Inverse Document Frequency (IDF) is statistically analyzed on the stems of all the extracted keywords by NLTK.}~(e.g., \textit{``is''}, \textit{``have''}) from the extracted verbs. 
    Intuitively, the more stylistic keywords of some style a story has, the more consistent it is with that style. Therefore, we propose to compare the numbers of keywords for different styles for style annotation.
    
    \paragraph{Normalized Numbers of Keywords}  Let $N_s$ denote the number of keywords for style $s$ in a story. We assume $N_s$ is a random variable, and follows a Gaussian distribution $\mathcal{N}(\mu_s, \sigma^2_s)$, where $\mu_s$ and $\sigma_s$ are the mean and standard deviation computed on the training set. Given a story which contains $n_s$ keywords for style $s$, we normalize $n_s$ to $n'_s=P(N_s\leqslant n_s)\in[0,1]$ for fair comparison between keywords for different styles.

        \begin{table}[!htbp]
    \small
    \centering
    \begin{tabular}{cccc}
    \toprule
    \textbf{Styles}&\textbf{Training}&\textbf{Validation}&\textbf{Test}\\
    \midrule
    \textbf{Emotion-driven}&17.7\%&18.0\%&17.9\%\\
    \midrule
    \textbf{Event-driven}&17.6\%&17.0\%&17.5\%\\
    \midrule
    \textbf{Others}&64.7\%&65.0\%&64.6\%\\
    \bottomrule
    \end{tabular}
    \caption{Distribution of stories annotated with different style for the training/validation/test set.}
    \label{tab:data_dist}
    \end{table}
    
    \paragraph{Annotation} 
    We annotate the style label $l$ for a given story by comparing its $n'_\text{emo}$ and $n'_\text{eve}$, which refer to the normalized numbers of keywords for emotion-driven and event-driven styles, respectively. 
    We annotate the story with $\langle \texttt{emo}\rangle$ if $n'_\text{emo}$ is higher than $n'_\text{eve}$, and $\langle \texttt{eve}\rangle$ otherwise.
    However, if both $n'_\text{emo}$ and $n'_\text{eve}$ are lower than $\tau_1$, or $|n'_\text{emo} - n'_\text{eve}| < \tau_2$, we annotate the story with $\langle\texttt{other}\rangle$ since there is no significant tendency to any styles. 
    $\tau_1$ and $\tau_2$ are hyper-parameters, which are set to 0.7 and 0.3, respectively.
    For stories labeled with $\langle\texttt{other}\rangle$, we select five words as the stylistic keywords from those keywords for emotion-driven and event-driven styles. 
    \autoref{tab:data_dist} shows the stylistic distribution of the dataset.

 


  \subsection{Baselines and Experiment Settings}
  We compare our model with GPT-2~\cite{radford2019language} and BART~\cite{lewis2020bart} as baselines. We fine-tune the baselines on ROCStories with the style tokens and the beginnings as input.

     We build our model based on BART. 
     Our approach can easily adapt to other pre-trained models such as BERT. 
    We set the scale factor in \autoref{eq:loss} to 0.2.
    For all models, We generate stories using top-k sampling~\cite{fan2018hierarchical} with $k=50$ and a softmax temperature of 0.8.

\subsection{Automatic Evaluation}
\paragraph{Evaluation Metrics}  \label{subsection:metrics}
  We use the following metrics for automatic evaluation:
    \textbf{(1) Perplexity (PPL)}. Since the automatically annotated  style labels may contain innate bias, we do not calculate the perplexity conditioned on the annotated styles for the stories in the test set. Instead, we calculate the perplexity of a model for each sample conditioned on two styles~(emotion-driven and event-driven), respectively, and then get the perplexity on the entire test set by averaging the smaller perplexity for each sample.
    \textbf{(2) BLEU~(B-n)} \citep{papineni2002bleu}\textbf{:} The metric evaluates $n$-gram overlap~($n=1,2$). For each beginning in the test set, we generate two stories conditioned on two styles, respectively. Then we calculate the BLEU score on the test set by averaging the higher BLEU with the reference story for each sample.
    \textbf{(3) Distinct~(D-n)}~\citep{li2016diversity}\textbf{:}  The metric measures the generation diversity with 
    the percentage of unique $n$-grams~($n=1,2$).
    \textbf{(4) Numbers of Stylistic Keywords (Number):} 
    We use the average $n'$~(described in \S\ref{sec:anno}) to evaluate how many consistent stylistic keywords the generated stories have. \textbf{(5) Lexical Style Consistency (LSC):} We calculate the percentage of the stories annotated with the consistent style in all generated stories using the annotation strategy described in \S\ref{sec:anno}.
    \textbf{(6) Semantic Style Consistency (SSC):} 
    It is a learnable automatic metric~\cite{DBLP:conf/emnlp/GuanH20}. We fine-tune BERT$_{\rm BASE}$ on the training set as a classifier to distinguish whether a story is emotion-driven, event-driven, or others with the automatic labels as the golden truth. For each style, we regard the average classification score on the style to measure the style consistency. \autoref{tab:ssc} shows the accuracy and F1-Scores of the BERT model on the test set. 

    \begin{table}[htp]
    \small
    \centering
    \begin{tabular}{c|ccc}
    \toprule
    \multirow{2}{*}{\textbf{Accuracy}} & \multicolumn{3}{c}{\textbf{F1-Score}} \\
                        & \textbf{Emotion-Driven}  & \textbf{Event-Driven}& \textbf{Other}     \\ \midrule
    89.7\% & 0.863                     & 0.838            & 0.922 \\
    \bottomrule
    \end{tabular}
    \caption{Accuracy and F1-Scores for each class of the BERT used in SSC.}
    \label{tab:ssc}
    \end{table}
    
    

\paragraph{Results}

We show the evaluation results of PPL and BLEU in \autoref{tab:all-res}. Note that we do not provide PPL for GPT-2 since it does not adopt the same vocabulary used in BART. 
We can see that our model has lower perplexity and higher word overlap with the human-written stories than baselines.

\begin{table}[!htbp]
\scriptsize
\centering
\begin{tabular}{cccc}
\toprule
\textbf{Models} & \textbf{PPL} $\downarrow$ & \textbf{B-1} $\uparrow$  & \textbf{B-2} $\uparrow$  \\ \midrule
\textbf{GPT-2} & N/A & 32.8 & 16.1   \\
\textbf{BART}  & 11.72 & 33.2 & 16.6 \\
\midrule
\textbf{Ours}  & \textbf{11.29}& \textbf{33.8} & \textbf{17.1} \\ \bottomrule
\end{tabular}
\caption{Automatic evaluation results on the entire test set. The Best results are highlighted in \textbf{bold}. $\downarrow$/$\uparrow$ indicates the lower/higher, the better.}
\label{tab:all-res}
\end{table}

We present the results of diversity and style consistency on the generated stories with different specified styles in  \autoref{tab:experiment-res}. Our model achieves comparable diversity with baselines, generates more keywords of the specified styles, and outperforms baselines in both lexical and semantic style consistency by a large margin. 

\begin{table}[!htp]
\scriptsize
    \centering
    \begin{tabular}{cccccc}
    \toprule
    \textbf{Models} &  \textbf{D-1} $\uparrow$  & \textbf{D-2 }$\uparrow$ & \textbf{Number} $\uparrow$ &\textbf{LSC}$\uparrow$& \textbf{SSC} $\uparrow$       \\ 
    \midrule
    \midrule
    \multicolumn{3}{l}{\textbf{Emotion-driven Style}}\\
    \midrule
    \textbf{GPT-2} &  0.679          & 0.924          & 0.454                    &0.243& 0.201                    \\
      \textbf{BART}  &  \textbf{0.701} & \textbf{0.952} & 0.538                  &0.366  & 0.298               \\
     \textbf{Ours}  & 0.697          & \textbf{0.952} & \textbf{0.623}&\textbf{0.474 }          & \textbf{0.371} \\ 
     \midrule
     \midrule
     \multicolumn{3}{l}{\textbf{Event-driven Style}}\\
     \midrule
    \textbf{GPT-2} &  0.675          & 0.925          & 0.375                    &0.107 & 0.088          \\
   \textbf{BART}  &  0.697          & 0.954          & 0.460                    &0.162& 0.129                   \\
            \textbf{Ours}  &  \textbf{0.698} & \textbf{0.955} & \textbf{0.591} &\textbf{0.309}          & \textbf{0.293}\\
    \bottomrule
    \end{tabular}
    \caption{Automatic evaluation results. The best results are highlighted in \textbf{bold}. 
    }
    \label{tab:experiment-res}
\end{table}
\begin{table*}[!htp]
\centering
\resizebox{\textwidth}{!}
{
\begin{tabular}{c|c|llll|llll|llll}
\toprule
\multirow{2}{*}{\textbf{Styles}}&\multirow{2}{*}{\textbf{Models}} & \multicolumn{4}{c|}{\textbf{Fluency}}             & \multicolumn{4}{c|}{\textbf{Coherence}}           & \multicolumn{4}{c}{\textbf{Style Consistency}}   \\ 
&                       & \textbf{Win}{(\%)} & \textbf{Lose}(\%) & \textbf{Tie}(\%) & $\kappa$ & \textbf{Win}(\%) & \textbf{Lose}(\%) & \textbf{Tie}(\%) & $\kappa$ & \textbf{Win}(\%) & \textbf{Lose}(\%) & \textbf{Tie}(\%) & $\kappa$ \\
\midrule
\multirow{2}{*}{\textbf{Emotion-driven}}&\textbf{Ours} vs. \textbf{GPT-2}          & 37.0**  & 19.0     & 44.0    & 0.751    & 52.0**  & 25.0     & 23.0    & 0.803    & 39.0*   & 22.0     & 39.0    & 0.773    \\
&\textbf{Ours} vs. \textbf{BART}           & 51.0**  & 22.0     & 27.0    & 0.672    & 52.0**  & 16.0     & 32.0    & 0.591    & 50.0*   & 31.0     & 19.0    & 0.640    \\
\midrule
\multirow{2}{*}{\textbf{Event-driven}}&\textbf{Ours} vs. \textbf{GPT-2}          & 40.0*   & 25.0     & 35.0    & 0.754    & 50.0*   & 30.0     & 20.0    & 0.735    & 54.0**  & 28.0     & 18.0    & 0.726    \\
&\textbf{Ours} vs. \textbf{BART }          & 43.0*   & 25.0     & 32.0    & 0.644    & 45.0**  & 22.0     & 33.0    & 0.635    & 48.0*   & 33.0     & 19.0    & 0.612    \\ \bottomrule
\end{tabular}
}
\caption{Manual evaluation results. The scores indicates the percentage of \textit{win}, \textit{lose} or \textit{tie} when comparing our model with a baseline. $\kappa$ denotes Fleiss' Kappa~\cite{fleiss1971measuring} to measure the inter-annotator agreement. * and ** mean p-value$<$0.05 and p-value$<$0.01 (Wilcoxon signed-rank test), respectively.}
\label{tab:human}
\end{table*}

    \subsection{Manual Evaluation}
    We conduct a pairwise comparison between our model and baselines. We randomly generate 100 stories from the test set for each style and model.
    For each pair of stories (one by ours, and the other by a baseline), we hire three annotators to give a preference~(\textit{win}, \textit{lose} and \textit{tie}) in terms of fluency, coherence, and style consistency. We adopt majority voting to make the final decisions among the annotators. We resort to Amazon Mechanical Turk for manual annotation. 
    As shown  in \autoref{tab:human}, all the results show moderate ~($0.4\leqslant\kappa\leqslant0.6$) or substantial~($0.6\leqslant\kappa\leqslant0.8$) agreement, and our model outperforms baselines significantly in fluency, coherence, and style consistency.

\begin{table*}[!htp]
\small
    \centering
    \begin{tabular}{cp{400pt}}
    \toprule
    \multicolumn{2}{l}{\textbf{Leading Context}: Bob has a girl friend.} \\ 
    \midrule
    \midrule
    \multicolumn{2}{l}{\textbf{Emotion-driven Style}}\\
    \midrule
    \textbf{GPT-2} & She wants to \event{take} a trip to Hawaii. She \event{goes} on vacation. She is at the beach. It is a \emotion{pretty} day.
            \\
    \midrule
      \textbf{BART}  &  One day, Bob \event{saw} a \emotion{cute} necklace on the sidewalk. Bob decided to \event{buy} it.  After buying it, Bob \emotion{loves} it.  Bob \emotion{likes} the necklace.\\
      \midrule
     \textbf{Ours}  & He is really \emotion{nervous} about her feeling around her. His girlfriend is very \emotion{protective}. Bob gets along \emotion{great} with her. Bob has a \emotion{wonderful} time with his girlfriend.\\ 
     \midrule
     \midrule
     \multicolumn{2}{l}{\textbf{Event-driven Style}}\\
     \midrule
    \textbf{GPT-2} &She \emotion{likes} her hair. She \event{takes} a few pictures of her friend's hair. He \event{takes} a picture of her hair and \event{posts} it. She \emotion{likes} it very much and she is \emotion{happy}.   \\
    \midrule
   \textbf{BART}  & He knew he was always going to be \emotion{mean} to her. After a while Bob \event{realized} that he was being \emotion{annoying}. He had to \event{leave} his job and \event{walk}. Now he has a new girlfriend and a new job.\\
   \midrule
    \textbf{Ours}&He has been \event{talking} to her all day. She stopped \event{listening} to him now. One day, she \event{says} his name and \event{walked} away. He decided to \event{break} up with her in another place.\\
    \bottomrule
    \end{tabular}
    \caption{Generated stories by different models with different specified styles. \emotion{Emotion-related} keywords and \event{event-related} keywords are highlighted in italic and bold, respectively. } 
    \label{tab:case-study}
\end{table*}

\section{Case Study}
\label{section:case_study}

\autoref{tab:case-study} shows several generated cases. We generate the stories using different models given the same leading context and specified style. 
For the \textit{emotion-driven} style, our model can generate various emotional keywords~(e.g., \textit{``nervous''}, \textit{``protective''}, \textit{``great''}, and \textit{``wonderful''}) and focus more on shaping the characters' personality. 
For the \textit{event-driven} style, 
our model can generate fluent stories with a reasonable event sequence. 
In comparison, baselines tend to confuse the two styles. For example, the stories generated by the baselines for the event-driven style still contain many emotional keywords~(e.g., \textit{``likes''}, \textit{``annoying''}). Besides, for the emotion-driven style,  the baselines generate fewer and repetitive emotional keywords. 
Furthermore, the baselines may suffer from more severe repetition~(e.g., ``take a picture'') than our model. 
And the baselines sometimes mix up or neglect some characters (e.g., GPT-2 and BART only cover one of \textit{``Bob''} and \textit{``his girlfriend''} but neglect the other one for the \textit{emotion-driven} style).
In summary, our model can generate more coherent stories with specified styles than the baselines.

\section{Conclusion}
    We present a pilot study on a new task, stylized story generation. We define story \textit{style} with respect to \textit{emotion} and \textit{event}, and propose a generation model which conditions on planned stylistic keywords. Comparative experiments with strong baselines show the promising results of the proposed model. Our work can inspire further research in this new direction.

\section*{Acknowledgements}
    This work was jointly supported by the NSFC projects (Key project with No. 61936010 and regular project with No. 61876096), and the Guoqiang Institute of Tsinghua University, with Grant No. 2019GQG1. We thank THUNUS NExT Joint-Lab for the support. 

\bibliographystyle{acl_natbib}
\bibliography{acl2021}

\appendix
\section{Implementation Details}
\subsection{Vocabulary}
The initial vocabulary of GPT-2/BART contains 50,258/50,265 tokens, respectively.
We add three style tokens ($\langle \texttt{emo}\rangle$, $\langle \texttt{eve}\rangle$,  $\langle \texttt{other}\rangle$) and three name tokens~($\langle \texttt{MALE}\rangle$, $\langle \texttt{FEMALE}\rangle$, $\langle \texttt{NEUTRAL}\rangle$) to the vocabulary. Therefore, the final vocabulary for GPT-2/BART/our model contains 50,264/50,271/50,271 tokens, respectively.
\subsection{Hyper-parameters}
We follow BART$_{\rm BASE}$'s hyper-parameters and initialize our model with the public checkpoint of BART$_{\rm BASE}$\footnote{\url{https://huggingface.co/facebook/bart-base/tree/main}}. 
Both the encoder and decoder contain 6 hidden layers with 768-dimensional hidden states. GPT-2$_{\rm BASE}$\footnote{\url{https://huggingface.co/gpt2/tree/main}} uses a 12-layer decoder with 768-dimensional hidden states. The batch size is 32 for all the models when training. We uses the AdamW optimization~\cite{DBLP:conf/iclr/LoshchilovH19} and the initial learning rate is $5\times 10^{-5}$. At inference time, we set the maximum sequence length to 120 tokens. 
    
    


\begin{table}[!hbt]
\centering
\begin{tabular}{@{}cccc@{}}
\toprule
\textbf{Models}         & \textbf{GPT-2}  & \textbf{BART}   & \textbf{Ours}   \\ \midrule
\textbf{Training Time} & 242min & 128min & 336min \\ \bottomrule
\end{tabular}
\caption{Training time for models in the experiments}
\label{tab:train_time}
\end{table}
\subsection{Runtime}
The runtime of fine-tuning of each model is reported in \autoref{tab:train_time}. We do the experiments on one GeForce GTX TITAN X GPU.

\section{Manual Evaluation}
\begin{figure*}[htbp]
    \centering
    \includegraphics[width=.83\textwidth]{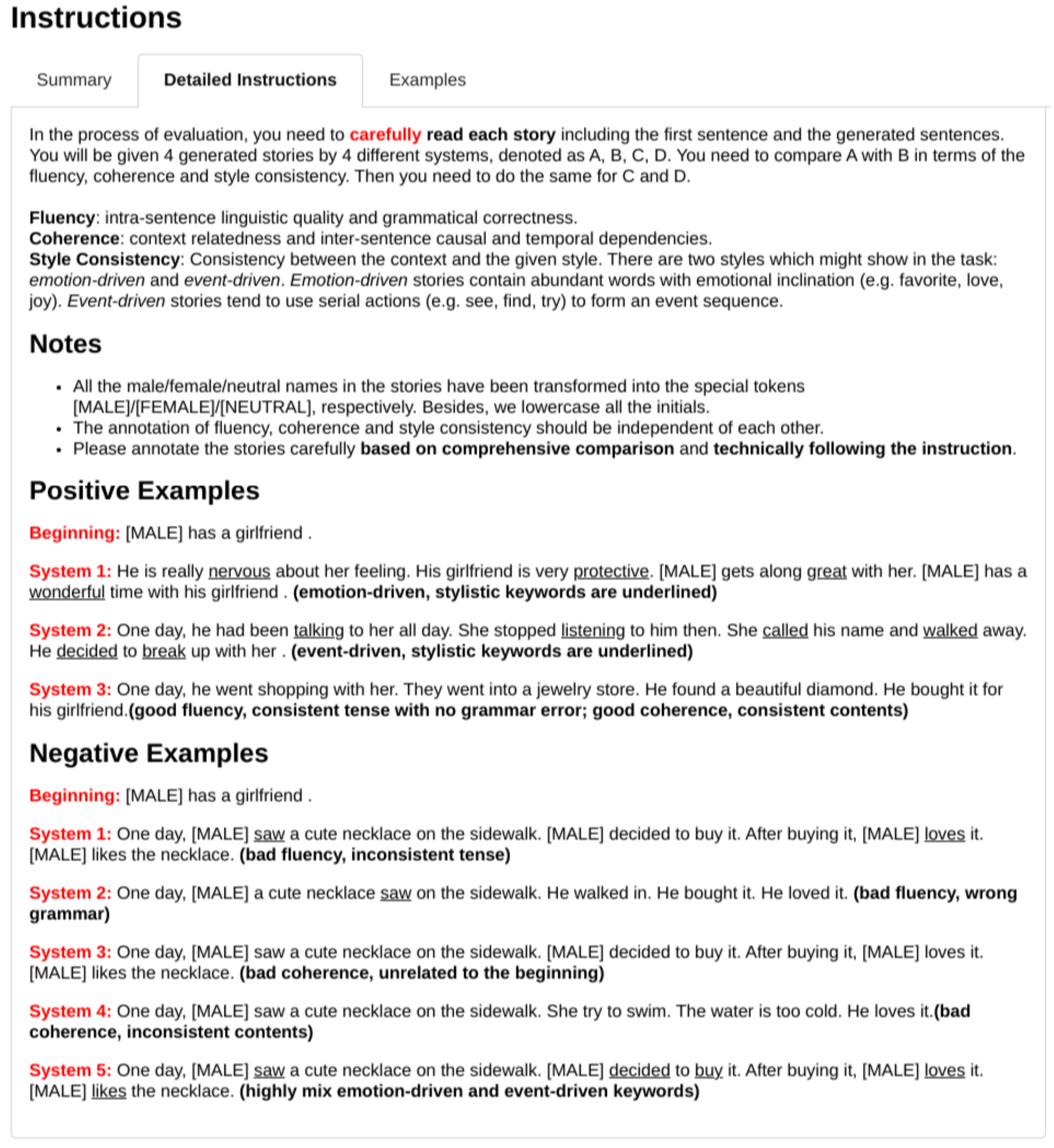}
    \includegraphics[width=.83\textwidth]{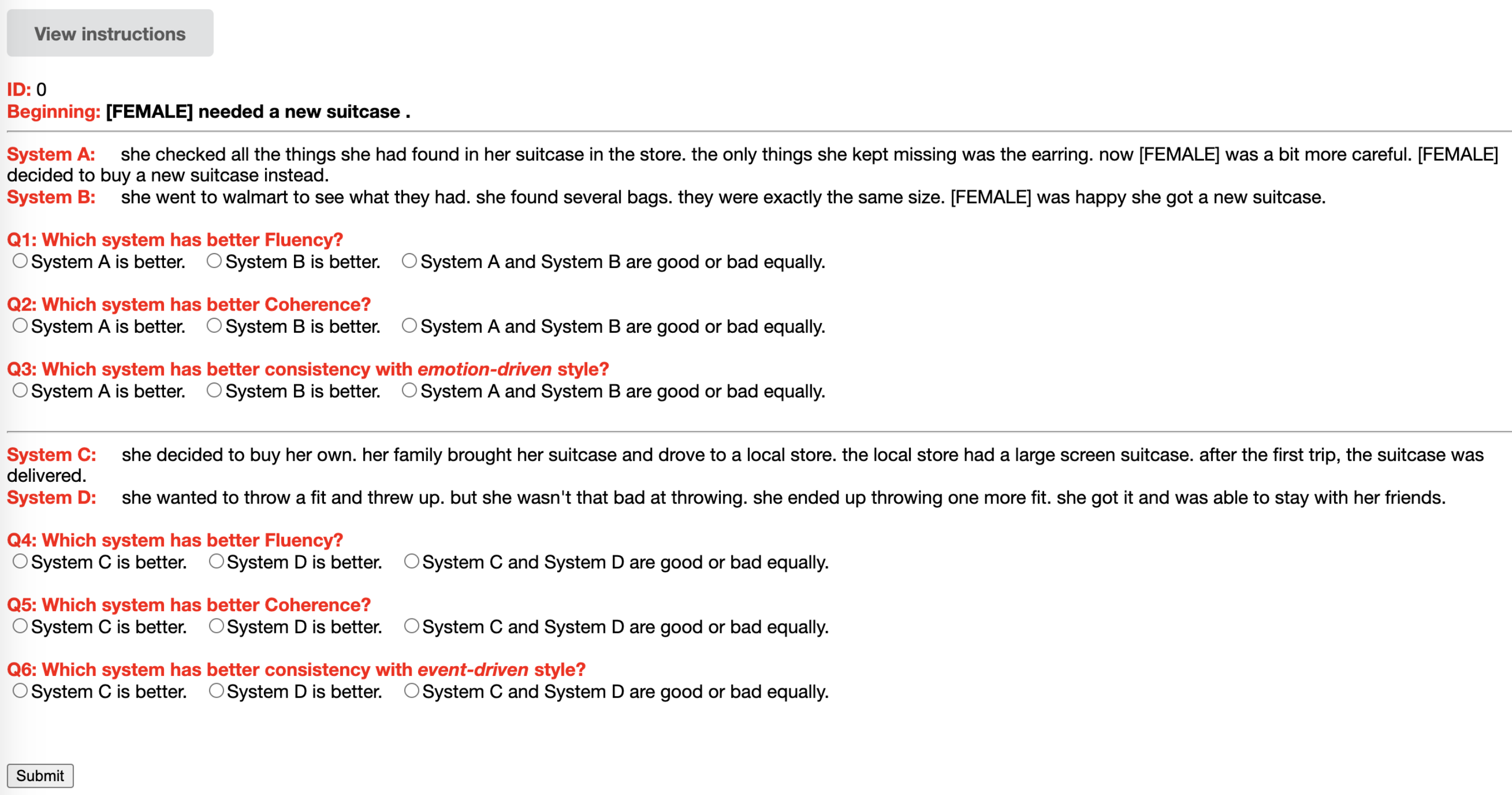}
    \caption{A screenshot of manual evaluation on AMT}
    \label{fig:amt_ex}
\end{figure*}

As described in the main paper, we conduct manual evaluation on AMT.
\autoref{fig:amt_ex} shows a screenshot of an annotation example on AMT.

\end{document}